# Definitions and Semantic Simulations
# Based on Object-Oriented Analysis and Modeling

Robert B. Allen

[0000-0002-4059-2587]

rba@boballen.info

**Abstract.** We have proposed going beyond traditional ontologies to use rich semantics implemented in programming languages for modeling. In this paper, we discuss the application of executable semantic models to two examples, first a structured definition of a waterfall and second the cardiopulmonary system. We examine the components of these models and the way those components interact. Ultimately, such models should provide the basis for direct representation.

**Keywords:** Dynamic Models, Model-Level, Multiscale Semantic Modeling, Rich Semantics, Scale-yoked Entities

## 1. Introduction

An ontology describes entities and relationships among entities. We go beyond typical ontologies to implement a model layer. Our approach develops executable semantic simulations using a programming language. While we take upper ontologies as providing a form of data typing [7], semantic modeling requires many additional features such as complex objects with states and mechanisms. In previous work, we have suggested the value of object-oriented analysis with rich semantics to the systematic description of natural systems. This is supported by Galton and Mizoguchi's definition of objects:

> …we identify an object as an interface between those processes which are internal and those processes which are external to it… ([15] p. 1)

The main goal of this work is the integration of many different approaches into a comprehensive framework. Our primary interest is in description rather than inference. Sections 2 and 3 provide examples of semantic modeling that are discussed in greater detail in later sections. Section 2 presents a structured definition of a waterfall. Section 3 provides an executable model of the cardiopulmonary system. Section 4 describes Parts of Objects while Section 5 considers their Behavior and States. Section 6 discusses Functions, Mechanisms, Systems, and Microworlds. Section 7 describes Modeling. Finally, Section 8 considers validation and extrapolating mechanisms as well as semantic modeling for scholarly communication.

## 2. Structured Definitions

### 2.1 Waterfall Example

The philosopher Heraclitus famously said: "You can never step into the same river twice." He concluded that "all is change". By comparison, other philosophers emphasize the static nature of objects. Still others adopt a hybrid that is also consistent with natural language, which has both verbs and nouns. In most cases, verbs act on nouns. When one is running, one is changing position in space. Nonetheless, Heraclitus's river is still widely discussed. The action of flowing is integral to the essence of what it means to be a river. A version of this quandary focusing on waterfalls has recently been discussed [15].

According to dictionary definitions, a waterfall is:

> … a cascade of water falling from a height, formed when a river or stream flows over a precipice or steep incline[1]

---

[1] https://www.lexico.com/en/definition/waterfall

and,

> an area where water flows over a vertical drop or a series of steep drops in the course of a stream or river.[2]

```
class StreamPath():
  def TraverseUpperStreamBed(self,i):
    for L in range(0,self.upperBedLength):
      self.water[i].X = self.water[i].X+10
      self.water[i].Y = self.water[i].Y-1
      self.water[i].Location="upper"
  def TraverseDrop(self,i):
    for V in range(0,self.verticalDrop):
      self.water[i].X = self.water[i].X+1
      self.water[i].Y = self.water[i].Y-10
      self.water[i].Location="drop"
  def Pool(self,i):
    self.water[i].Location="pool"
  def __init__(self, water):
    self.water = water
    self.upperBedLength=1000
    self.verticalDrop=100

class WaterPortion():
  X = 0
  Y = 0
  Location="null"

class Waterfall():
  def WaterFlowing(self):
    i=0
    while(True):
      self.water.append(WaterPortion())
      self.bed.TraverseUpperStreamBed(i)
      self.bed.TraverseDrop(i)
      self.bed.Pool(i)
      print(i,self.water[i].Location)
      i = i+1
  def __init__(self):
    self.water = []
    self.bed=StreamPath(self.water)
    self.WaterFlowing()
def run():
  W = Waterfall()
```

*Figure 1: Python program that defines and implements a simple simulated waterfall.*

Figure 1 shows a structured definition for a waterfall that covers the key features of the definitions. We have incorporated specific parameters, so it is executable and provides a simple simulation of a waterfall. There is a class for the Waterfall itself and two other classes for the main parts of a waterfall: the StreamPath and the Water. In the program, the water flows in the stream bed with a small slope and then comes to a steep drop. Because of the constructor ( __init__ ), the water starts to flow as soon as the program is initialized. The water is modeled as individual Portions (Sections 2.1, 4.2) and is assigned to different location states as it flows (e.g., upper stream, drop, pool).

---

[2] https://en.wikipedia.org/wiki/Waterfall

## 2.2 Commentary about the Waterfall Example

Defining objects with object-oriented classes highlights and provides a structure for resolving the issues raised by philosophers and lexicographers. The ongoing transitions of portions, or droplets, of water are integral to the definition of rivers and waterfalls. Similarly, transitions are integral to some other objects such as stars and systems such as the solar system, but most objects are initially static and demonstrate changes only when they are the subject of Transitionals (e.g., a person starts to run). There are many refinements we could introduce. If it were a stream rather than a river, we could adjust the amount of water accordingly and model the bed in greater detail. We could also model the waterfall if it froze. Each cycle of the while loop could check the current state of the water to make sure it is fluid.[3]

Coordination between molecules and macroscopic effects is a common issue for semantic modeling of fluids (Sections 4.2, 8.1). In Figure 1, we model the movements of small Portions of water. However, an actual portion of fluid would not retain continuity over time. Thus, this is an idealized model in which the Portions remain unified.

Because we are dealing with a qualitative, or nominal, model we could have simply moved Portions across the qualitative states. Rather, to add fidelity to the model, we moved Portions across numeric units. This makes it an ordinal model rather than a purely qualitative model (see Section 7.1).[4] Ideally, extending model resolution would be relatively seamless. Because this is an interpreted Python program, we could pause it as it runs and manually examine the value of any variable. In the future, we could add a graphic control panel to support more flexible ways to interact with the model (Section 7.3).

## 2.3 Frame Net Analysis

Frame Net [2, 28] is a linguistic resource based on the theory of frame semantics [13]. The key idea is that rather than just activating the meaning of scattered words in a sentence, a broad context is activated. Frame Net identifies and describes approximately 1500 Frames. One of those frames is Fluidic_Motion:

> A Fluid moves from a Source to a Goal along a Path or within an Area[5]

The core Frame Elements of Fluidic_Motion are Fluid, Source, Goal, and Path/Area. Other, non-core Frame Elements include Configuration, which specifies parameters such as the volume and speed of the Fluid.

In addition, specific Lexical Entries (i.e., words) are associated with each Frame. Flowing is a Lexical Entry associated with the Fluidic_Motion Frame from which it inherits the Frame Elements. Although Frame Net does not give a detailed analysis of Flowing it does provide this definition:

> To move with a continual change of place among the constituent particles[6]

The WaterFlowing routine in Figure 1 embeds the core Frame Elements along with features of this definition. Alternatively, the Fluidic_Motion Frame Elements could have been represented with the familiar functional notation:

> Flow (Fluid, Source, Pool, Path)

---

[3] This is a version of two-phase validation (Section 8.1). Validation could also be addressed by contracts such as those implemented in the Eiffel programming language. For semantic modeling, the contracts could include inheritance and semantic tests.

[4] Ordinal models are not full quantitative models. Quantitative models might employ massively parallel computation using the Multiscale Object-Oriented Simulation Environment (MOOSE) [12, 14]. For the waterfall, a full quantitative model might be based at the molecular level and also fully model the erosion of the stream bed. Nonetheless, although the simulation can be highly numerical, the description may still be mostly qualitative.

[5] https://framenet2.icsi.berkeley.edu/fnReports/data/frameIndex.xml?frame=Fluidic_motion

[6] https://www.merriam-webster.com/dictionary/flow

In the case of the waterfall, a Path is a recursive series of segments with width, length, and slope. Other Parameters could be incorporated by adding the Configuration non-core Frame Element. In addition, the flexibility of the code could be enhanced if we defined an abstract class FluidicMotion.

In Frame Net, the Lexical Entry for a Waterfall is associated with the Natural_Features Frame:

> The Locale is a geographical location as defined by shape. This frame includes natural geographic features, including land/ice forms and bodies of water.

Potentially, we could develop a microworld terrain model for the surrounding environment (Section 6.4) by linking the Natural_Features Frame and other Frames with the Fluidic_Motion Frame.

A much larger standardized vocabulary should be developed that could support hooks for the interoperability of the frames. Beyond a collection of structured Frame Elements, there could be standardized computational descriptions for the Lexical Units. In addition, for some definitions, functionality is as important as structure, and those programmatic definitions could include that aspect (Section 6.1).

# 3. Executable Semantic Model of the Cardiopulmonary System

## 3.1 Overview

We also developed a stylized, textbook-level executable semantic model of the Cardiopulmonary System as a proof-of-concept for our approach. This example highlights the application of semantic modeling to Mechanisms and Systems. As shown in Figure 2, $O_2$-rich air moves from the nose to alveoli in the lungs. There, $O_2$ diffuses into the blood and $CO_2$ from the blood passes back into the lungs. The $O_2$ then circulates through the heart and out to the body. Cells in the body use the $O_2$ and return $CO_2$ into the blood. In addition, a sensor in the medulla triggers the diaphragm if there is too much $CO_2$ in the blood. Our model is asynchronous and has several threads. We implemented two main subsystems: circulation and respiration.

For the circulatory sub-system, the contraction of the heart spreads across its four chambers. Portions of blood are moved throughout the system with each heartbeat. We modeled capillaries at the lungs, at a typical body cell, and at a sensor in the medulla. For the respiratory sub-system, a portion of $O_2$-rich air is drawn through the nose into the alveoli in the lungs. The $O_2$ diffuses into a portion of blood in the capillaries and, at the same time, $CO_2$ passes from the blood into the portion of air. That portion of air is then exhaled through the nose out to the external air. For the capillaries, we modeled a physical structure (i.e., the capillary tube), a non-material spatial region, and a small portion of fluid (either blood or air) contained within that spatial region. As with the Portions of Water in the Waterfall example, we assume that these Portions pass through the circulatory system intact (Sections 2.2, 4.2).

The modeled blood and air flows follow the solid black lines shown in Figure 2. For both the respiratory and circulatory systems, explicit connection relationships were set between the nodes. Further, the presence of a connection was validated before any Portions were pushed. Because the network never changed in this version of the model, that check was mostly redundant. Rather than trying to synchronize pushing individual portions through the circulatory system, all updates were calculated and then implemented simultaneously. The timing for the modeled motion of the fluids was rough. The respiratory system was triggered independently by the Medulla and the circulatory system by the SA Node.

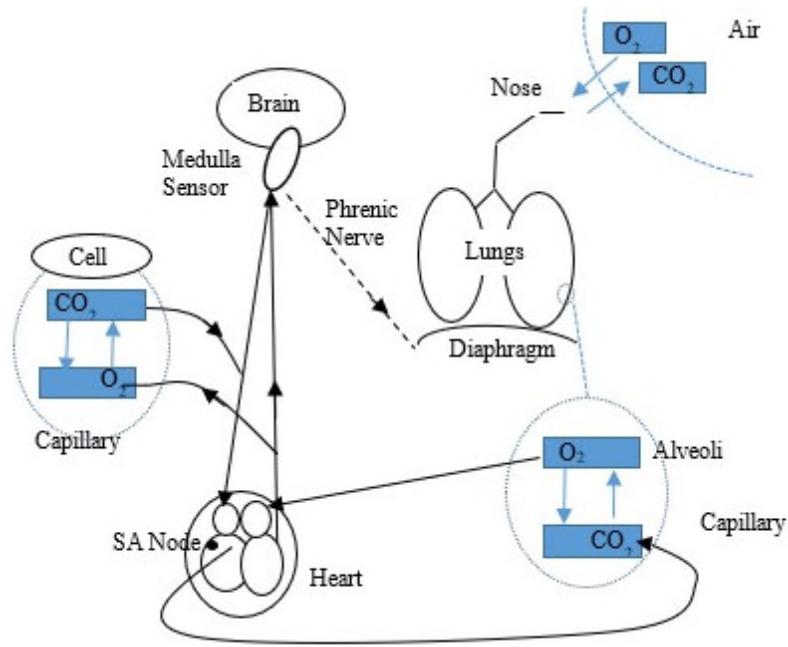

*Figure 2: A schematic model of the cardiopulmonary system. The major organs (lungs, heart, and brain) are shown. The solid black lines connecting the organs show the path for blood and air flow. The black dotted line shows the Phrenic Nerve which controls the diaphragm. Also shown are enlargements of the capillaries at a typical cell and at a typical alveolus. Respiratory and circulatory sub-systems can be identified along with mechanisms for nerve signaling and gaseous diffusion. While the connectivity is included in the model, relative spatial positions are not.*

Figure 3 shows a sample output as the program runs. Because the systems are asynchronous, the updates are interleaved. In the figure, we can see examples where the SA Node triggers and Portions of blood are pushed in the circulatory sub-system, cases in which diffusion occurred, and an example of a contraction of the diaphragm causing a breath to be taken.

```
CellCapBlood O2 diffusion          mixing external air
pushed LeftAtriumBlood             diffusion check
pushed LeftVentricleBlood          AlvCapBlood O2 diffusion
pushed MedullaCapBlood             CellCapBlood O2 diffusion
pushed cellCapBlood                pushed LeftAtriumBlood
pushed RightAtriumBlood            pushed LeftVentricleBlood
pushed RightVentricleBlood         pushed MedullaCapBlood
pushed AlvCapBlood                 pushed cellCapBlood
trigger updates                    pushed RightAtriumBlood
SAnode pulse                       pushed RightVentricleBlood
inhale cycle                       pushed AlvCapBlood
past phrenicNerve trigger          trigger updates
into diaphram contract             diffusion check
completed inhale ExternalAir to Nose Air   CellCapBlood O2 diffusion
completed inhale Nose Air to Alv Air       mixing external air
```

*Figure 3: Two panels showing a section of continuous output from the cardiopulmonary semantic simulation system.*

### 3.2 Commentary about the Cardiopulmonary System Example

This model implements systems, sub-systems, and mechanisms. The fluid motion of the air and the blood could be implemented with the FluidicMotion abstract class described above in Section 2.2. The model incorporates many simplifications; it uses only simple motion and simple structures.

The nerve impulse was not modeled in detail but could have been implemented as sodium channels opening and closing along the neuron. The sensors can be seen as providing information about the $CO_2$ level, which is transmitted to the diaphragm.[7]

We could generate a family of related models of different levels of complexity. On the one hand, we could model cell metabolism. On the other hand, we could extend this model to include interaction with other body systems such as the digestive system. Moreover, the model of the entire cardiopulmonary system as described here could be adapted for use across species of mammals.

## 4. Parts

### 4.1 Functional and Structural Parts

We focus here on Functional Parts, which are Objects that are essential for a state change in the model (Section 6.1). Mechanisms link behavior and parts; they show how the functionality is accomplished. We also allow Structural Parts such as the stream bed in Section 2. Structural Parts are essential for the physical instantiation of a Mechanism without contributing directly to the functionality.[8] Structural Parts support the interaction of parts but do not show state changes themselves. For instance, the bracket holding a car's carburetor in place would be a structural part supporting the carburetor. The bracket may be composed of several separate pieces and the array may be considered an Assembly.[9]

Several fields have developed resources for structural and functional parts that could be incorporated into our models. For instance. mechanical engineers and computer graphics researchers have developed CSG (Constructive Solid Geometry)[10] which is a structural modeling framework that allows for the relative motion of parts such as the motion of joints. Similarly, frameworks from anatomy (e.g., [10, 17]) and structural biology could be incorporated.

### 4.2 Composition and Portions of Matter

Beyond the usual notion of Functional Parts (e.g., the engine, tires, and body of a car), we sometimes consider the materials of which an Object is composed. We saw examples in Sections 2 and 3 where water, blood, and air are better described as Substances rather than Objects. [18] describes a typography of Substances and Mixtures. In Sections 2 and 3, we defined Portions of the Substances and took the Portions as Objects [31]. Some of the challenges in this area can be traced to the uneven treatment of atoms and molecules as Objects. Adding or subtracting a few molecules of water does not materially change a lake. The difference between water molecules and the lake is a granular perspective [18, 31]. We propose a multi-granular perspective which allows modeling the different levels at the same time.

## 5. States, Transitions, and Behavior

Objects change; but how should we model those changes? We adopt State Transitions. The notion of State is widely recognized in information processing. Roughly, a State is a distinct condition of an object for a given period of time. We distinguish between Object-Centered States, which are associated with a single Object, and Ensemble States, which describe the interaction of several independent Objects.

---

[7] The mechanism for transmitting the nerve signal is relatively straightforward. The notion of "information" may be most useful as a shortcut at the model-level when the mechanisms are not specified.

[8] [24] proposes that the Structure of an Object consists of all the internal relationships of that Object. Further, he proposes that Structure should be considered as an entity in its own right. This seems consistent with the definition of Objects by classes in an object-oriented programming language but differs somewhat from the use of Structure in this section.

[9] Although such a bracket can be said to provide a function -- that of positioning the carburetor to allow it to interact with other parts –that is a secondary function.

[10] https://en.wikipedia.org/wiki/Constructive_solid_geometry

### 5.1 Object-Centered States

The State of an Object is interwoven with its Parts. In many cases, it is the activity of Parts that makes a State. One example is that when a person is running their legs are moving quickly back and forth. Another example would be describing phase changes in a material which is due to the excitation of its atoms/molecules. State changes for a traffic light (cf., [8]) can be modeled as State changes of separate (green, yellow, red) lamps within the light. But, ultimately the State of those lamps is due to the activity of electricity (i.e., electrons) flowing. The example of the Waterfall (Section 2) is unusual because the flowing water is not a State but is integral to it being a waterfall. We refer to the traffic light and the waterfall as having multi-granular states because the parts (e.g., the electrons) are at a different level than the parent object. Other State changes are due to changes in functionality (e.g., usable/unusable) although these differences may, ultimately, also be based on changes of Parts. Still other types of State changes are due to changes in space and/or time.

### 5.2 Ensemble Interaction and States

We can model the States of interacting Objects. An underlying question is when independent Objects may better be considered as a single, unified Object. Examples include one and the clothes one is wearing, a couple who gets married, or a car that gets a new coat of paint. These might be implemented as Relational Qualities connecting two Objects (e.g., [9] p 97-98) though it would be helpful to have more nuanced descriptions of the possible types of relationships that could be used. In some cases, we can treat the new, unified object as a replacement for the previous object. The car with a new coat of paint could be considered as an update to the car. In other cases, new entities exist but are highly transient such as a party with its revelers or a chemical bond existing fleetingly during a reaction. In still other cases, these Objects retain a context-dependent duality. When a person becomes part of a couple, that person's professional activities may be relatively unchanged while other activities are as part of the couple. This is comparable to the relationship between molecules and the atoms that compose them. In yet other cases, the Objects retain their distinct identities but are joined by their participation in an activity. We consider a pianist as distinct from the piano that is being played although both are involved in the same scenario.

### 5.3 Transitionals and Behavior

States imply the possibility of State changes. We use the more general term Transitionals to include other types of transformations of objects such as birth, death, splits, and merges. State changes are often changes of parts, but not all changes of parts result in State changes of the parent Object or vice versa (Section 5.1). Thus, we can distinguish Behavior from State changes. When a person is running, their legs are constantly moving but are not constantly changing State. Likewise, water molecules are always vibrating and spinning but small changes in those activities do not change the State of the water.

In some cases, we might ask whether an Object has States or even Behavior. For example, some physical objects such as a bridge or chair provide a function (i.e., a river crossing or a seat), but have little apparent motion. Nonetheless, they do deform when used and, presumably, the bonds between their atoms and molecules are affected. At a low level of granularity, they are in a state of tension when providing a function.

### 5.4 Relationship to Object-Oriented Analysis and Modeling

In earlier papers, we have noted the similarity of semantic modeling to object-oriented analysis and modeling. There is not unanimity about the definition of object-oriented modeling, but we can consider several of the most common features associated with it - abstraction, inheritance, encapsulation, and polymorphism (e.g., [20]).

An example of abstraction is our use of abstract methods to describe the contraction of the heart or diaphragm. Objects in our models exhibit inheritance; for instance, different mammals could inherit a general cardiopulmonary system model such as described in Section 3.[11] Encapsulation suggests that the methods used by an Object should be hidden and inaccessible except by hooks at the top level. However, because we have multi-granular modeling, the Parts of Objects and the methods associated with them are generally exposed (Section 5.1). Polymorphism means that methods may be overloaded. While our examples do not include Polymorphism, potentially they could. Object-oriented languages such as Smalltalk propose that objects should communicate via message passing. In the cardiopulmonary example, the Phrenic Nerve can be thought of as passing messages to the diaphragm. For models of physical interaction with collisions between objects, those collisions can be considered as a type of message passing.

# 6. Functions, Causation, Mechanisms, Systems, and Microworlds

## 6.1 Functions

Mechanical engineers often identify three dimensions for describing their models: Function, Behavior, and Structure (FBS or SBF) [16]. In Section 4 we addressed Structure such as Parts and Assemblies and in Section 5 we addressed Behavior (and State). In this section, we turn to Function. We include how something is used along with its function.

To assert that an Object has a Function we need to know its broader context. If someone in the Middle Ages had, somehow, created an object identical to a modern carburetor, we would not say that that object has the function of a carburetor. In other words, we assert that a Function is relative to a Mechanism, System, or Scenario, and a description of a Function needs to include a context – perhaps by identifying a Mechanism or System with which it is associated.[12] For example, while the study of anatomy is most often concerned with structure, reconciling parts and functions is a traditional issue for anatomy [10, 22, 26].

## 6.2 Causation

Modeling sequences of events assumes that the events do not happen at random, that they are, in some sense, based on earlier events. We say that events are caused by those earlier events. Nonetheless, the notion of causation is highly contentious. Much of the problem is due to the difficulty of ascertaining causation retrospectively. That often involves possibly unreliable evidence and uncertain inference. Similarly, assessing causation is difficult when the Objects, Parts, and Transitions are ambiguous. However, in a model, where we define all of the Objects, Parts, and Transitions, there is little controversy about causation.

Basic science identifies types of Objects in the world as well as their Parts and Transitions. In other words, basic science can be seen as developing consistent descriptive frameworks ([9] pp 12-13). By comparison, applied science attempts to use those frameworks to predict or explain real-world phenomena. Given science's success, it is reasonable to identify objects as entities with "causal powers" [24] and that what is real is what is consistent with scientific results.

---

[11] It is unlikely that simple single inheritance can always be applied across entire complex objects and systems. Presumably, some way of indicating exceptions and inheritance by Function/Use or Behavior could be adopted.

[12] Perhaps the need for the existence of a Mechanism or System is what the Basic Formal Ontology (BFO) intends by declaring that Functions are Realizables. For example, [9] (p 104) states that "to detoxify its containing organism is a function of this liver" ([9] p 104). However, BFO has no clear notion of Mechanism, System, or Scenario.

### 6.3 Mechanisms

[4] described some simple semantic Mechanisms based on Petri Nets.[13] Mechanisms based on Petri Nets are the foundation of functional descriptions. Although we can trace a mechanism between any two causally connected Objects, we most often focus on Mechanisms that are directly responsible for the functioning of a system. In some cases, such as the Central Dogma in biology, which describes the steps from the transcription of DNA to the production of proteins, a Mechanism may be multi-granular. That is, a transition at a low level can dramatically affect the dynamics of a system at a higher level across time.

A side effect is a change that is not directly needed for the completion of a given Mechanism. For instance, a reaction may produce heat that does not affect the completion of the Mechanism that produced it. But that heat may affect other Mechanisms. Although developers are urged to avoid them in well-designed systems, side effects are inherent in many natural systems and must be included in models.

### 6.4 Systems, Sub-Systems, Microworlds, and Scenarios

A system is a complex object which is composed of several interacting Mechanisms (cf., [2]). Several different types of systems may be identified. Many are purely feed-forward, while others have feedback. Some of the latter have an internal regulator and manage homeostasis while others of them are chaotic even if they attempt regulation. We can model much of the behavior of systems with feedback with semantic tools, although for purely semantic models the non-linearities need to be described qualitatively.

"A Microworld is a restricted, idealized model of the world containing only those relations and entities of interest in the particular reasoning system being designed" ([11] p 6). A System implemented in a Microworld may show spatial relationships (e.g. "next to") and may reflect other constraints (e.g., "is symmetrical"). We also allow ambient properties such as temperature, humidity, and even gravity.[14]

To distinguish the Microworld as a frame or platform from its representation of the world, we call the latter Scenarios. We have considered Scenarios at several different points in this work. A terrain model incorporating the Waterfall example and the Cardiopulmonary system in Figure 2 are Scenarios. We might use a variation of Figure 2 to describe what happens when a person is running or what happens when a person's heart stops beating. This is not intended to be generalized inference about broken mechanisms (see Section 8.1); rather, it is a targeted extension of a model to handle a specific situation. It can be considered as a semantic modeling alternative to referential ontologies [23].

## 7. Modeling

### 7.1 Model Operators

In addition to the data types such as Objects, Transitionals, and Mechanisms, we also allow familiar data modeling features such as cardinality. Thus, we may specify that a mammal has two ears, two eyes, and two or four legs. Because the programming language is integral to the modeling, we also need to recognize language features such as looping, conditionals, concurrency, and threading.

While the current models are primarily qualitative (i.e., categorical or nominal), they also include simple binary comparisons (e.g., high and low concentrations of $O_2$ and $CO_2$). They could be extended across richer levels recognized by data analysis as ordinal, interval, and ratio models.

---

[13] In comparison to the model of the cardiopulmonary system (Section 3), the models in [4] were ad hoc functional accounts without systematic descriptions of Behavior or Structure.
[14] Standard temperature and pressure (STP) is the default.

### 7.2 Meta Operators and Annotation

Meta-operators are elements incorporated into a model to make it more tractable. Along with Annotations, they show how the model does not fully reflect reality. The meta operators would include approximations to continuous operations (e.g., diffusion), cross-granular descriptions [21], approximations to stochastic processes, picking typical or random examples, and compensation for non-linearities. Structured annotations may include describing simplifications (i.e., eliminating unnecessary detail) and idealizations (i.e., intentionally inaccurate statements which, nonetheless, improve clarity).[15] Other types of annotations could include comments from users.

### 7.3 Explanations and User Interaction

As described in Section 2.2, a graphical control panel could be developed for user interaction with the programs. The interface would have modes suitable for different types of users (e.g., students, researchers, editors). In addition to graphical interaction, there could be a natural language interface. That could employ text generation based on the model operators and annotations described in the previous sections. The text generation could also apply discourse elements [1] such as those from Rhetorical Structure Theory (RST) [19]. It might even support tutoring. We may also be able to develop a grammar for manipulating the model. Such a grammar could be a sort of interactive, model-based programming language.

## 8. Discussion

### 8.1 Validation and Extrapolating Mechanisms

While our primary goal is description, it is useful to validate the programs as much as possible. For instance, we should validate that the application of Transitionals does not violate any of the assertions made by relationships. The example in Section 2 includes some checking, but more is needed. This can be done with two-phase validation which was described in our previous work [6]. That is, applying traditional validation of triples after every step of the model.

### 8.2 Semantic Modeling for Scholarly Communication

We expect that models like those described here will be the basis of direct representation research reports [3]. Such reports would describe scientific research based on a structured knowledge base. In addition to modeling natural phenomena, the models would incorporate structured descriptions of research workflows and data analysis workflows [4]. Claims could be made about aspects of Mechanisms and evidence from observations applied to support or contradict those claims. Deemphasizing text will eventually lead to less need for text mining.

Potentially, both students and researchers would benefit from browsing knowledge structures. Standards for constructing the models could be developed and libraries of models could be collected. If we develop libraries of composable mechanisms[16] and models, then users should be able to bridge out to related models and also to drill down from more general models to more detailed models. Moreover, broken Objects and Transitions will likely affect the Mechanisms in which they participate (e.g., [25, 26, 29, 30]). A well-documented library of Mechanisms would be of great value for making predictions about the impact of and solutions to those broken Object and Transitions.

---

[15] See https://plato.stanford.edu/entries/models-science/
[16] For example, the introduction to many research papers starts by listing known mechanisms associated with the phenomenon under study.

### 8.3 Envoi

We have sketched a broad framework and emphasized a broad synthesis of many approaches. In Section 2 we developed a programmatic implementation of natural language definitions which is consistent with Frame Net. In Section 3, we implemented a semantic simulation of interacting components of the cardiopulmonary system. In Sections 4-7, we considered issues for the implementation. Although these examples are relatively simple, they are potentially composable and scalable. We envision that the current generation of ontologies will eventually be replaced by terms with fully-structured definitions. However, challenges remain, such as how to manage consistency in the assumptions made when describing the interaction of several related mechanisms.